# Optimized SC-F-LOAM: Optimized Fast LiDAR Odometry and Mapping Using Scan Context

Lizhou Liao, Chunyun Fu[*], Binbin Feng, and Tian Su

*Abstract*— LiDAR odometry can achieve accurate vehicle pose estimation for short driving range or in small-scale environments, but for long driving range or in large-scale environments, the accuracy deteriorates as a result of cumulative estimation errors. This drawback necessitates the inclusion of loop closure detection in a SLAM framework to suppress the adverse effects of cumulative errors. To improve the accuracy of pose estimation, we propose a new LiDAR-based SLAM method which uses F-LOAM as LiDAR odometry, Scan Context for loop closure detection, and GTSAM for global optimization. In our approach, an adaptive distance threshold (instead of a fixed threshold) is employed for loop closure detection, which achieves more accurate loop closure detection results. Besides, a feature-based matching method is used in our approach to compute vehicle pose transformations between loop closure point cloud pairs, instead of using the raw point cloud obtained by the LiDAR sensor, which significantly reduces the computation time. The KITTI dataset is used for verifications of our method, and the experimental results demonstrate that the proposed method outperforms typical LiDAR odometry/SLAM methods in the literature. Our code is made publicly available for the benefit of the community[1].

## I. INTRODUCTION

Simultaneous localization and mapping (SLAM) plays an important role in the fields of robotics and autonomous driving; it also lays the foundation for path planning and motion control of robots and driverless vehicles [1]. According to different sensors used, SLAM solutions can be generally classified into two major types: LiDAR SLAM and visual SLAM. Compared with cameras, LiDAR sensors are advantageous in measurement accuracy, measurement range, and resistance to environmental interference, which makes LiDAR SLAM generally more accurate in localization [2].

### A. Literature Review

LiDAR SLAM has undergone more than 30 years' development. Up to now, a large number of LiDAR SLAM-related works have been published in the literature. Solutions to SLAM problems have also evolved from statistical-based approaches in early years [3] to graph optimization methods nowadays [4]. In the last decade, typical

This work was supported by the Graduate Research and Innovation Foundation of Chongqing, China under Grant CYS22015. *(corresponding author: Chunyun Fu)*

Lizhou Liao, Binbin Feng and Tian Su are with the College of Mechanical and Vehicle Engineering, Chongqing University, Chongqing 400044, China. E-mail: {lizolizhou, fengbinbin, sutian}@cqu.edu.cn.

Chunyun Fu is with the College of Mechanical and Vehicle Engineering, Chongqing University, Chongqing 400044, China. E-mail: fuchunyun@cqu.edu.cn.

[1] https://github.com/SlamCabbage/Optimized-SC-F-LOAM

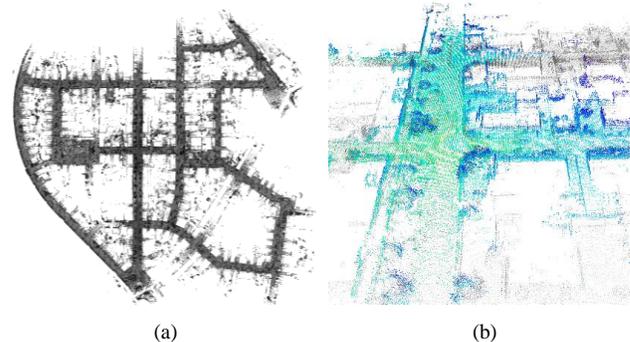

Fig. 1: Example of the proposed method on KITTI dataset. (a) shows the mapping effect of the proposed method on KITTI dataset sequence 00. (b) shows an instance of loop closure detection, where the colored part is the point cloud in the current frame and the gray part is the historical map.

LiDAR SLAM solutions have developed into a framework containing three major parts, including front-end odometry, loop closure detection and global optimization [5-7].

Frame matching generally includes raw point cloud matching and feature-based matching. For raw point cloud matching, the most classical method is Iterative Closest Point (ICP) [8] which finds the closest point in Euclidean space for each point in the other point cloud, iteratively minimizes this distance residual and converges to the final estimated pose. However, ICP requires costly computation due to the huge number of points. Another common method for raw point cloud matching is the Normal Distribution Transformation (NDT) [9] which puts the point cloud data into grids and then calculates the local normal distribution in each grid, so as to reduce the computational cost of iteration by point-to-normal distribution matching. However, the matching accuracy with NDT is dependent on grid size, and it deteriorates in case of large grids. Zhang and Singh [10] proposed a feature-based inter-frame matching method known as Lidar Odometry and Mapping (LOAM), which has become a benchmark inter-frame matching method up to now. In this method, feature points (edge points and plane points) are extracted based on local point smoothness, and the feature points extracted from the current frame are matched with corresponding established feature submap to obtain the pose transformation. Based on Ceres Slover [11], Wang et al. [2] proposed Fast LiDAR Odometry and Mapping (F-LOAM), which further optimizes the frame matching accuracy of LOAM and reduces the computational cost. The pose estimation accuracy of F-LOAM is ranked among the best according to the KITTI odometry benchmark [12]; yet presence of cumulative error due to its lack of loop closure detection reduces its effectiveness in large-range estimation scenarios.

As for loop closure detection, a number of vision-based solutions have been proposed in the literature, such as [13-15]. These loop detection methods are subject to many limitations as a result of poor robustness to ambient illumination [16]. LiDAR-based loop closure detection methods can be categorized broadly into local descriptor methods and global descriptor methods. The former ones relate to calculating the location signatures of detected key points, constructing a bag-of-words model, and finally matching the location signatures with those of other frames to obtain the similarity between them [17-20]. This category of methods has unstable performance in matching local key points in outdoor scenarios where unstructured objects (e.g., trees) are present and in cases where point cloud density varies with distance. To solve this problem, global descriptor methods were subsequently proposed in the literature. Kim and Kim [21] put forward a global descriptor method to reduce the dimensionality of a frame of point cloud data and store such data in a two-dimensional matrix. Dube et al. [22] proposed a segmentation-based scene recognition method, known as SegMatch. This SegMatch method has been used as the loop closure detection module in [23, 24] to achieve global optimization of vehicle pose. However, additional point cloud processing is required in [23, 24], which increases the computational cost of loop closure detection.

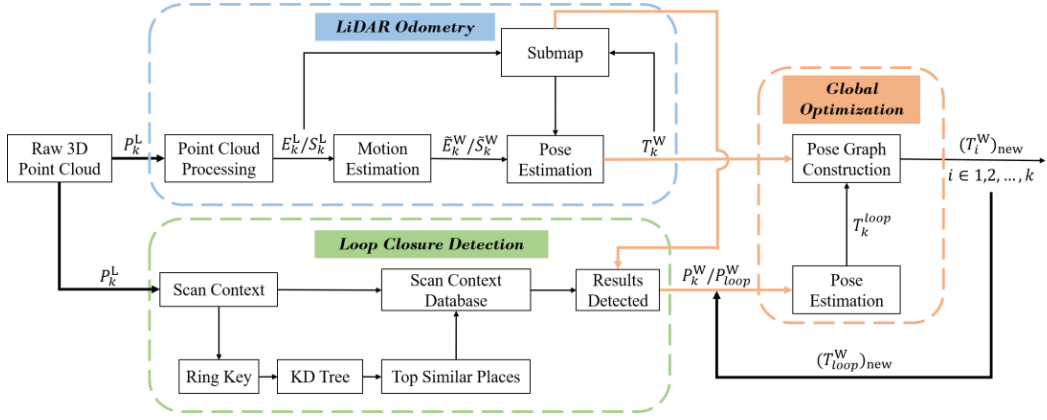

Fig. 2. Schematic of the Simple-SC-F-LOAM. The three different-colored dashed boxes represent three parallel processing modules. The blue dashed box represents the LiDAR odometry module, the green one represents the loop closure detection module, and the orange one represents the global optimization module.

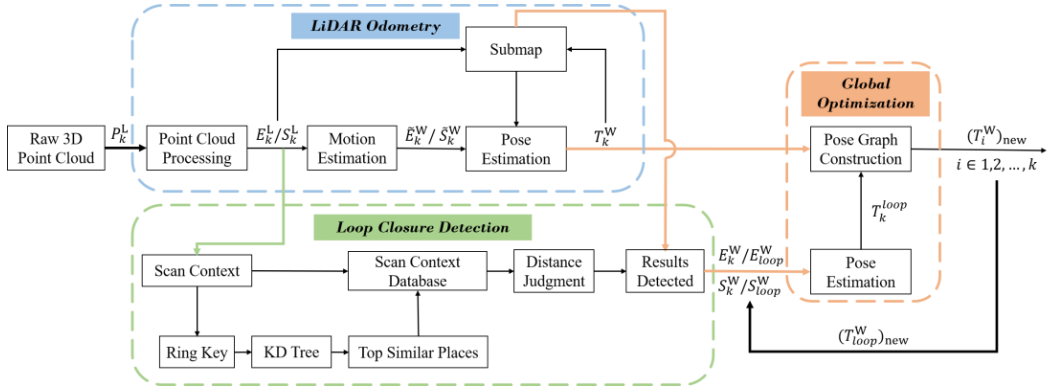

Fig. 3. Schematic of the proposed Optimized-SC-F-LOAM. The three different-colored dashed boxes represent three parallel processing modules. The blue one represents the LiDAR odometry module, the green one represents the loop closure detection module, and the orange one represents the global optimization module.

For global optimization, the most common approach is pose graph optimization [25]. Several popular open source C++ libraries [11, 26, 27] can be used to solve global optimization problems, among which GTSAM [26] archives good optimization result at low computational cost [28]. For this reason, GTSAM is used in the present study for global optimization.

B. Problem statement

In the existing literature, many LiDAR SLAM methods have only front-end odometry without loop closure detection [2, 10, 29-32]. As a result, vehicle/robot pose estimation errors accumulate as the explored scene expands. Although some LiDAR SLAM methods in the recent literature [23, 24, 33] have incorporated loop closure detection, most of them constructed pose constraints by simply including one loop closure method in the SLAM framework, which does not necessarily guarantee optimal localization accuracy.

SC-LeGO-LOAM, a combination of LeGO-LOAM [30] and Scan Context [21], is able to achieve global optimization of vehicle pose upon loop closure and thereby reduce the accumulated errors. However, the ICP algorithm [8] adopted in SC-LeGO-LOAM for computing the pose transformation between loop closure point cloud pairs requires costly computation and may affect real-time establishment of loop constraints. F-LOAM [2] is a lightweight LiDAR odometry method with greatly reduced computational cost, but it is unable to achieve accurate localization in large scenarios due to lack of loop closure detection that can reduce the cumulative pose estimation error. Therefore, it is highly necessary to improve integration of LiDAR odometry and loop closure detection, mainly because 1) reducing the cumulative error in pose estimation leads to improved SLAM accuracy; 2) lowering the cost of calculating pose transformation between loop closure point cloud pairs lightens the workload of the SLAM method.

*C. Original Contribution*

To solve the above problems, a lightweight LiDAR SLAM method, i.e. Optimized Fast LiDAR Odometry and Mapping Using Scan Context (Optimized-SC-F-LOAM), is proposed in this paper, based on F-LOAM [2] and Scan Context [21]. The Optimized-SC-F-LOAM method comprises three parts: LiDAR odometry, loop closure detection and global optimization; it can be used to achieve more accurate localization at less computational cost. Main contributions of this paper are as follows:

1. Feature-based matching method is used, instead of the conventional raw point cloud matching (such as ICP), to calculate the pose transformation between loop closure point cloud pairs, which significantly reduces the computational cost.

2. An adaptive distance threshold is employed to identify whether a loop closure is established. This threshold reduces the possibility of false loop closure detection and further improves the localization accuracy. Moreover, due to reduced false loop closure detections, the computation time of global optimization is also shortened.

3. The entire SLAM framework is made publicly available for the benefit of the community under appropriate license agreement.

*D. Outline of Paper*

The remainder of the paper is organized as follows: Section II introduces the method for comparison in this paper; Section III explains the proposed method in detail, including LiDAR odometry, loop closure detection and global optimization; Section IV provides the comparative experimental results and discussions on results; Section V summarizes the whole paper.

## II. BACKGROUND

In this paper, we name the direct combination of F-LOAM [2] and Scan Context [21] as Simple-SC-F-LOAM. The specific framework of Simple-SC-F-LOAM is shown in Fig. 2. In this framework, the 3D raw point cloud $p_k^L$ is passed into the LiDAR Odometry module and the Loop Closure Detection module to derive the vehicle pose $T_k^W$ in the current frame relative to the global coordinate system, as well as the loop closure detection result. If a loop closure is detected, the pose $T_{loop}^W$ resulting from the LiDAR odometry is used to transform the loop closure point cloud pair to the same coordinate system, leading to $p_k^W$ and $p_{loop}^W$. Then, the vehicle pose transformation $T_k^{loop}$ between the point cloud pair $p_k^W$ and $p_{loop}^W$ is calculated using the ICP algorithm, and it is added to the pose graph together with the odometry pose $T_k^W$. Lastly, global optimization is performed to obtain the optimized vehicle pose.

Real tests have proved that the method proposed in this work takes merely 28% of the time required on average by Simple-SC-F-LOAM to compute the pose transformation between a loop closure point cloud pair. Besides, the original Scan Context method results in many false loop closure detections that not only increase the computational cost but also reduce the accuracy of final vehicle localization. To tackle this false detection problem, the proposed method is designed with an adaptive distance threshold that can eliminate most false loop closure detections resulting from Scan Context, thereby improving the accuracy of vehicle localization.

## III. METHODOLOGY

In this section, the proposed SLAM method is elaborated. The entire framework of the method proposed in this paper is shown in Fig. 3. Similar to that of Simple-SC-F-LOAM, the framework of the proposed method also consists of three modules: LiDAR Odometry, Loop Closure Detection and Global Optimization, but it differs in the following three aspects:

1. Feature point clouds obtained by the LiDAR odometry are used as the input to the Loop Closure Detection module.

2. An adaptive distance threshold is used to identify whether a loop closure is established, so as to reduce false loop closure detections.

3. Feature-based matching is employed to compute the vehicle pose transformation between a loop closure point cloud pair.

*A. LiDAR Odometry*

In the LiDAR Odometry module, a framework substantively consistent with [2] is employed. This LiDAR odometry mainly includes four sub-modules: 1) Point Cloud Processing, 2) Motion Estimation, 3) Pose Estimation and 4) Submap Update. These sub-modules are introduced one by one below.

1) Point Cloud Processing: This sub-module in the LiDAR odometry first receives the raw point cloud data collected by LiDAR, and uses the same method as described in [2, 10, 32] to extract feature points based on local point smoothness. A point with smoothness greater than the threshold is considered as an edge point, otherwise it is deemed as a plane point. Then, the edge point cloud $E_k^L$ and the plane point cloud $S_k^L$ are passed to the Motion Estimation sub-module and the Submap Update sub-module.

2) Motion Estimation: This sub-module in the LiDAR odometry receives the incoming edge point cloud $E_k^L$ and plane point cloud $S_k^L$, and uses the uniform motion model [2] to estimate the initial pose transformation from the LiDAR coordinate system in the current frame to the global coordinate system. By means of this initial pose transformation, the edge point cloud $E_k^L$ and the plane point $S_k^L$ are transformed into the global coordinate system to obtain $\tilde{E}_k^W$ and $\tilde{S}_k^W$, which are then passed into the Pose Estimation sub-module.

3) Pose Estimation: This sub-module receives the edge point cloud $\tilde{E}_k^W$ and the plane point cloud $\tilde{S}_k^W$ in the global coordinate system, and the feature submap passed in from the Submap Update sub-module. Firstly, 5 adjacent points corresponding to each feature point in the feature submap are identified. Then, the adjacent points of the edge feature points form corresponding feature lines, and the adjacent points of the plane feature points form corresponding feature surfaces. The sum of all distances from feature points to corresponding feature lines (or feature surfaces) forms a nonlinear equation. Secondly, the Gauss-Newton nonlinear optimization method is used to iteratively solve pose transformation $T_k^W$ from the current frame to the global coordinate system. For specific optimization steps, please refer to [2]. Lastly, $T_k^W$ is passed into the Submap Update sub-module to update the feature submap, and simultaneously into the Global Optimization module to build the pose graph.

4) Submap Update: This sub-module receives the feature point clouds $E_k^L$ and $S_k^L$ in the current frame, and the pose transformation $T_k^W$ from the LiDAR coordinate system in the current frame to the global coordinate system. Then, $E_k^L$ and $S_k^L$ are projected via $T_k^W$ to the global coordinate system and added to the feature map for use by the Pose Estimation sub-module in the next frame.

### B. Loop Closure Detection

The Loop Closure Detection module of the proposed method has incorporated the well-known Scan Context framework [21]. For brevity, the details of Scan Context are not repeated and we hereby introduce the major differences between the proposed method and the original Scan Context.

A loop closure detection framework usually constructs a global descriptor for detection using the input point clouds. Different from Scan Context which uses raw point cloud data to form the global descriptor, the proposed method constructs the global descriptor using the edge point cloud $E_k^L$ and the plane point cloud $S_k^L$ obtained from the LiDAR Odometry module. These two types of feature points, $E_k^L$ and $S_k^L$, contain almost all useful information in the point cloud raw data. The use of these feature points reduces the influence of clutters, improves the accuracy of loop closure detection, and shortens the loop closure detection time.

The original Scan Context method uses a fixed distance threshold to determine and exclude highly similar but far-away point cloud pairs. As a result, setting a large fixed distance threshold may cause many false loop closure detections. To overcome this shortcoming, in this study we have introduced an adaptive distance threshold to replace the conventional fixed distance threshold. By this means, false loop closure detections can be reduced. Once the loop closure detection results are obtained, we use the poses $T_k^W$ and $T_{loop}^W$ corresponding to the loop closure point cloud pair to calculate the distance between the two frames, as follows:

$$\tilde{T}_k^{loop} = T_{loop}^{W\ -1} T_k^W \quad (1)$$

$$d = \sqrt{\tilde{T}_k^{loop}.x^2 + \tilde{T}_k^{loop}.y^2 + \tilde{T}_k^{loop}.z^2} \quad (2)$$

where $\tilde{T}_k^{loop}$ represents the pose transformation between the current frame and the loop frame obtained through the LiDAR odometry results $T_k^W$ and $T_{loop}^W$, $\tilde{T}_k^{loop}.x$, $\tilde{T}_k^{loop}.y$, and $\tilde{T}_k^{loop}.z$ denote the $x$, $y$, and $z$ components of the vector connecting the preceding two frames (i.e. pointing from the current frame to the loop frame), and $d$ is the magnitude of this vector. If $d$ is greater than a certain threshold $d_{thre}$, then it is considered that no loop closure is detected.

Since the cumulative error increases with the distance traveled by the vehicle, we design the threshold $d_{thre}$ as a function of the number of odometry frames:

$$d_{thre} = 20 + k/n \quad (3)$$

where $k$ represents the number of keyframes, and $n$ is a design parameter dependent on the cumulative error of the odometry. This threshold can reduce false loop closure detections, thereby improving global optimization results and efficiency of the entire SLAM algorithm.

### C. Global Optimization

Reducing the cumulative error in vehicle pose estimation is a frequently discussed issue in the existing literature. The global optimization method used in this paper consists of two main sub-modules: 1) Pose Estimation and 2) Pose Graph Construction.

1) The Pose Estimation sub-module receives loop closure detection results from the Loop Closure Detection module, and based on such results, it obtains point clouds of the current frame and the loop frame in the global coordinate system from the LiDAR Odometry module. It should be noted that for computation of vehicle pose transformation between a loop closure point cloud pair, these two point cloud frames that constitute a loop closure must be in the same coordinate system [21]. For this purpose, we first project the current frame to the global coordinate system based on $(T_{loop}^W)_{new}$ (which will be explained in the next paragraph), and construct a submap of the loop-frame point cloud and its neighbors in the global coordinate system. Then, we use the pose estimation method in the LiDAR Odometry module to calculate the pose transformation $T_k^{loop}$ between the current-frame point cloud and the submap so constructed, and pass this pose transformation to the Pose Graph Construction sub-module for constructing the pose graph.

2) The Pose Graph Construction sub-module receives the pose transformation $T_k^{loop}$ passed in by the Pose Estimation sub-module and the pose transformation $T_k^W$ passed in by the LiDAR Odometry module. The received odometry pose $T_k^W$ is added to its corresponding pose graph node, and pose $T_k^{loop}$ is added as the edge corresponding to the node. Following that, the GTSAM algorithm [26] is used for global optimization of the pose graph to derive the optimized pose $(T_i^W)_{new}, i \in 1, 2, ..., k$.

## IV. EXPERIMENT EVALUATION

### A. Experiment Setup

To validate the effectiveness of our method in real scenes, we adopted sequences 00 and 05 from the KITTI dataset [34]

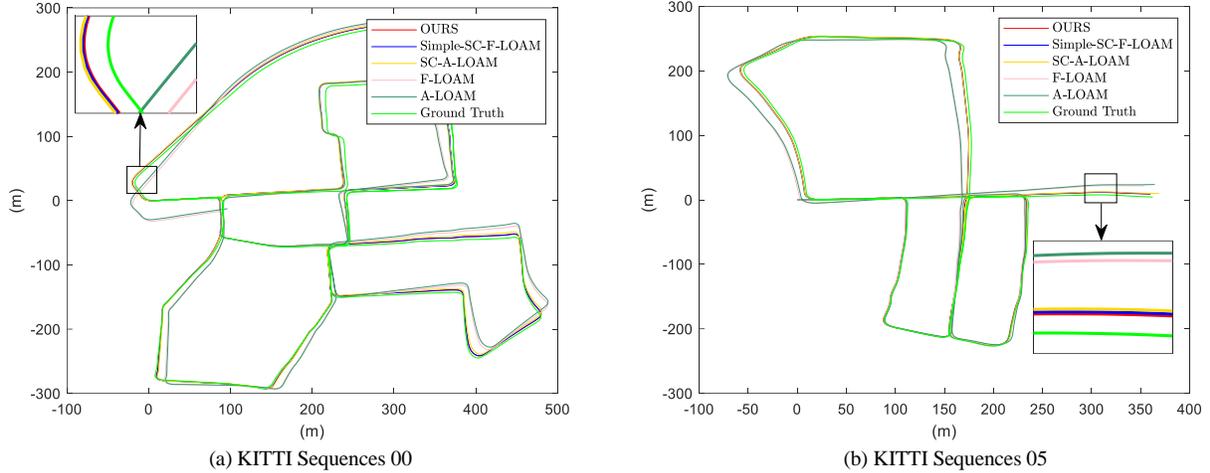

Fig. 4. Comparison of trajectories on KITTI dataset: (a) and (b) are the trajectories produced by all competing methods using KITTI sequences 00 and 05.

**TABLE I**. Results on KITTI Sequence 00 and Sequence 05.

| Methods | Datasets | ATE (%) | ARE (deg/100m) | Time (ms/loop frame) |
|---|---|---|---|---|
| Simple-SC-F-LOAM | 00 | 0.75 | 0.27 | 501 |
| | 05 | 0.51 | 0.24 | 422 |
| F-LOAM | 00 | 1.13 | 0.37 | - |
| | 05 | 0.81 | 0.32 | - |
| SC-A-LOAM | 00 | 0.83 | 0.34 | 444 |
| | 05 | 0.50 | 0.27 | 439 |
| A-LOAM | 00 | 1.09 | 0.40 | - |
| | 05 | 0.87 | 0.37 | - |
| LeGO-LOAM | 00 | 2.17 | 13.4 | - |
| | 05 | 1.28 | 0.74 | - |
| LOAM | 00 | 0.78 | - | - |
| | 05 | 0.57 | - | - |
| OURS | 00 | **0.75** | **0.27** | **135** |
| | 05 | **0.43** | **0.22** | **129** |

for evaluation. All tests were conducted based on the Robot Operating System (ROS), which was installed on a laptop equipped with an AMD R5-5600H CPU, a 16 GB RAM and the Ubuntu platform.

*B. Evaluation on KITTI dataset*

We used the open source KITTI dataset to test the performance of the proposed algorithm in gentle outdoor scenes. Since the proposed method is mainly intended to optimize pose estimation by LiDAR odometry under the condition of loop scenes, the sequences 00 and 05 in the KITTI dataset which contain loop closures were used to evaluate the localization accuracy and computational efficiency of the proposed method. For comparison purposes, the proposed method is compared against some typical LiDAR SLAM methods in the literature [2, 10, 21, 30] by using the same data sequences.

Fig. 4 shows the trajectories of the proposed method, the compared methods [2, 10, 21, 30] and the ground truths. We see that for both sequences, the proposed method outperforms the competing methods by providing trajectories closest to the ground truths. We also computed the Average Translational Error (ATE), the Average Rotational Error (ARE), and the computation time between loop closure point cloud pairs in these two sequences for each method. Table I shows that the proposed method has the smallest ATE and ARE for both sequences. At the same time, the proposed method also achieves the optimal computation time for both sequences. Specifically, for sequence 00, the time consumed by the proposed method is 73.1% and 69.6% less than those of Simple-SC-F-LOAM and SC-A-LOAM, respectively. As for sequence 05, the time consumed by the proposed method is 69.4% and 70.6% lower than those of Simple-SC-F-LOAM and SC-A-LOAM, respectively. Note that the ATE and ARE produced by LeGO-LOAM and LOAM were directly obtained from [31], and other results were computed by the authors using the source codes provided by the competing methods, on the same testing laptop as described above.

*C. Discussions on results*

To sum up, in gentle outdoor scenes, the proposed method has achieved great improvements in aspects of computation time and localization accuracy relative to the compared methods. There are two main reasons for these improvements. One is that feature point matching (instead of ICP matching) is used in the proposed method to calculate the vehicle pose transformation between loop closure point cloud pairs, thereby reducing the computation time. The other is that a loop judgment criterion is re-adaptively designed in the proposed method for loop closure detection with Scan Context to reduce false loop closure detections, so that a more accurate pose can be obtained after global optimization.

## V. Conclusion

In this paper, a computationally efficient LiDAR-based SLAM framework is proposed, which achieves better localization and mapping results at a lower computational cost. By combining a currently efficient LiDAR odometry method (F-LOAM) with a LiDAR loop closure detection method (Scan Context), the proposed method is designed with an adaptive distance threshold (instead of a fixed threshold) for loop closure detection, thereby optimizing loop closure detection performance. Such design has enabled the proposed method to achieve better localization accuracy and much higher computational efficiency.

In order to validate the effectiveness of the proposed method, we used the real scene dataset - KITTI dataset for comparative tests with several typical SLAM approaches in the literature. The results show that the proposed method outperforms the compared methods in terms of localization accuracy and computation time of pose transformation between loop closure point cloud pairs.